\documentclass[conference]{IEEEtran}
\IEEEoverridecommandlockouts

\usepackage{cite}
\usepackage{amsmath,amssymb,amsfonts}
\usepackage{algorithmic}
\usepackage{graphicx}
\usepackage{textcomp}
\usepackage{xcolor}
\usepackage{epsfig}
\usepackage{xcolor, soul}
\sethlcolor{red}
\usepackage{boldline}
\usepackage{amsmath}
\usepackage{amssymb}
\usepackage{soul}
\usepackage{url}
\usepackage{booktabs}
\usepackage{comment}
\usepackage{amsthm, nccmath}
\usepackage{mathtools, cases}
\usepackage{color}
\usepackage{lipsum}
\usepackage{float}
\usepackage{multirow}
\usepackage{xcolor}
\usepackage{hhline,colortbl}
\usepackage{ctable}
\usepackage{hyperref}
\usepackage{subcaption}
\usepackage{pifont}
\newcommand{\xmark}{\ding{55}}

\def\BibTeX{{\rm B\kern-.05em{\sc i\kern-.025em b}\kern-.08em
    T\kern-.1667em\lower.7ex\hbox{E}\kern-.125emX}}
\begin{document}


%

\title{Sharpening Lightweight Models for Generalized Polyp Segmentation:
A Boundary Guided Distillation from Foundation Models\\
\thanks{This work is supported by the Anusandhan National Research Foundation (ANRF), Government of India, under project number CRG/2023/007397 and ANRF/ARGM/2025/002890/TS.}
}

\author{\IEEEauthorblockN{Shivanshu Agnihotri}
\IEEEauthorblockA{\textit{Dept. of Computer Science and Engineering}
 \\
\textit{Malaviya National Institute of Technology}\\
Jaipur, India, 302017 \\
2024rcp9057@mnit.ac.in}
\and
\IEEEauthorblockN{Snehashis Majhi}
\IEEEauthorblockA{\textit{INRIA Sophia Antipolis}\\
\textit{ Côte d'Azur University}\\
France
\\
snehashis.majhi@inria.fr}
\and
\IEEEauthorblockN{Deepak Ranjan Nayak}
\IEEEauthorblockA{\textit{Dept. of Computer Science and Engineering}
 \\
\textit{Malaviya National Institute of Technology}\\
Jaipur, India, 302017\\
drnayak.cse@mnit.ac.in}
}

\maketitle

\begin{abstract}
Automated polyp segmentation is critical for early colorectal cancer detection and its prevention, yet remains challenging due to weak boundaries, large appearance variations, and limited annotated data. Lightweight segmentation models such as U-Net, U-Net++, and PraNet offer practical efficiency for clinical deployment but struggle to capture the rich semantic and structural cues required for accurate delineation of complex polyp regions. In contrast, large Vision Foundation Models (VFMs), including SAM, OneFormer, Mask2Former, and DINOv2, exhibit strong generalization but transfer poorly to polyp segmentation due to domain mismatch, insufficient boundary sensitivity, and high computational cost. To bridge this gap, we propose \textit{\textbf{LiteBounD}, a \underline{Li}gh\underline{t}w\underline{e}ight \underline{Boun}dary-guided \underline{D}istillation} framework that transfers complementary semantic and structural priors from multiple VFMs into compact segmentation backbones. LiteBounD introduces (i) a dual-path distillation mechanism that disentangles semantic and boundary-aware representations, (ii) a frequency-aware alignment strategy that supervises low-frequency global semantics and high-frequency boundary details separately, and (iii) a boundary-aware decoder that fuses multi-scale encoder features with distilled semantically rich boundary information for precise segmentation. Extensive experiments on both seen (Kvasir-SEG, CVC-ClinicDB) and unseen (ColonDB, CVC-300, ETIS) datasets demonstrate that LiteBounD consistently outperforms its lightweight baselines by a significant margin and achieves performance competitive with state-of-the-art methods, while maintaining the efficiency required for real-time clinical use. Our code is available at \href{https://github.com/lostinrepo/LiteBounD}{GitHub repository}.

\end{abstract}

\begin{IEEEkeywords}
Polyp Segmentation, Boundary-Guided Distillation, Foundation Models, High-Low Frequency Modulation 
\end{IEEEkeywords}

\section{Introduction}

Colorectal Cancer (CRC) remains a serious global health concern, ranking third among all cancer types and contributing significantly to cancer-related deaths. Most CRC cases originate from colorectal polyps- an abnormal tissue growth on the inner colon lining. Early detection and removal of such polyps is paramount for effective prevention \cite{morgan2023global}. Although colonoscopy is considered as the clinical gold standard for polyp detection, this procedure is highly operator-dependent and prone to inter-observer variability. Moreover, reported polyp miss rates of 6–27\% \cite{ahn2012miss} raise serious concerns, indicating the urgent need for accurate and reliable computer-aided polyp segmentation methods, thereby assisting clinicians in making precise interventions.

A plethora of encoder-decoder Convolutional Neural Network (CNN)-based models have been proposed for polyp segmentation over the past few years, including early architectures such as U-Net and its variants \cite{ronneberger2015u}, \cite{jha2019resunet++}. Despite achieving satisfactory performance, these early models often struggle to capture fine-grained boundary details. Following this, several architectures such as PraNet \cite{fan2020pranet}, MSNet \cite{zhao2021automatic}, SFA \cite{fang2019selective}, and M$^2$SNet \cite{zhao2023m}, have been introduced to enhance boundary awareness and handle scale variations of polyps. While these methods achieve notable improvements, their limited receptive fields hinder modeling of global contextual dependencies, leading to limited generalization. To this end, Vision Transformer (ViT)-based polyp segmentation models, such as CTNet \cite{xiao2024ctnet}, MCT-Net \cite{chakraborti2024mct}, PVT-Cascade \cite{rahman2023medical}, and Polyp-PVT \cite{Dong2023}, have been developed, owing to their ability to capture global relationships through self-attention, thereby resulting in remarkable performance improvements. However, such models typically incur high computational costs and demonstrate suboptimal generalization.

Despite these advances, inherent challenges, such as significant variations in polyp appearance and frequent occurrence of ambiguous or weak boundaries, continue to hinder robust polyp segmentation. Large-scale Vision Foundation Models (VFMs), including SAM \cite{kirillov2023segment}, CLIP \cite{radford2021learning}, OneFormer \cite{jain2023oneformer}, MaskFormer \cite{cheng2021per}, Mask2Former \cite{cheng2022masked}, and  DINOv2 \cite{oquab2024dinov}, have recently advanced segmentation by learning fine-grained visual representations with strong cross-domain generalization. However, their direct adaptation for polyp segmentation is constrained by insufficient domain-specific knowledge and substantial computational demands, hindering deployment in resource-constrained clinical environments. 
A recent work proposes SAM-Mamba \cite{dutta2025sam} to improve generalization via adapter-based tuning and the Mamba-Prior module. However, it remains computationally intensive (e.g., 103M parameters and 423 GFLOPs), limiting its applicability in real-time clinical settings. While lightweight encoder–decoder models such as U-Net remain a de facto choice for medical segmentation, they often exhibit limited robustness to low-contrast polyps and demonstrate poor generalization. To bridge the gap between generalization and efficiency, the recently introduced Polyp-DiFoM \cite{agnihotri2026sam} distills knowledge from multiple foundation models into a compact architecture, achieving strong generalization while maintaining efficiency. However, it still struggles to handle weak or ambiguous polyp boundaries effectively. these limitations highlight the necessity of a potential strategy that can effectively transfer the VFMs knowledge to lightweight models while preserving boundary-related cues.

To this end, we propose \textbf{LiteBounD} - a \underline{Li}gh\underline{t}w\underline{e}ight \underline{Boun}dary-guided \underline{D}istillation framework that transfers rich semantic and structural priors from VFMs into lightweight segmentation models. Additionally, LiteBounD employs a high-low frequency modulation strategy to decompose foundation model features into global and boundary-sensitive components, which are then distilled into lightweight baselines such as U-Net, U-Net++, and PraNet. This enables precise boundary refinement while preserving semantic coherence. Extensive experiments across five benchmark datasets---Kvasir-SEG, CVC-ClinicDB, ETIS, ColonDB, and CVC-300---demonstrate that LiteBounD consistently outperforms vanilla baselines, achieving superior accuracy and cross-dataset generalization under diverse imaging conditions.


In summary, our main contributions are as follows:
\begin{itemize}
    \item \textbf{Modular Distillation Framework:} We introduce LiteBounD, a plug-and-play distillation pipeline that infuses boundary-aware priors from VFMs (SAM, DINOv2, OneFormer) into lightweight segmentation models (U-Net, U-Net++ and PraNet), enabling efficient and scalable deployment.

    \item \textbf{Boundary-Aware Distillation:} We derive rich semantic and boundary-aware representations from foundation models that accurately discriminate polyp from non-polyp regions. These representations are subsequently decomposed into low- and high-frequency features using a high–low modulation strategy. The resulting features are then distilled into the baseline network, enabling precise refinement of structural details while simultaneously strengthening global semantic understanding.
    
    \item \textbf{Superior Generalization:} LiteBounD demonstrates consistent performance gains across five public datasets under seen and unseen conditions, validating its robustness and generalization capability in varied clinical scenarios.
    
\end{itemize}

\begin{figure*}[t]
  \centering

   \includegraphics[width=0.8\linewidth]{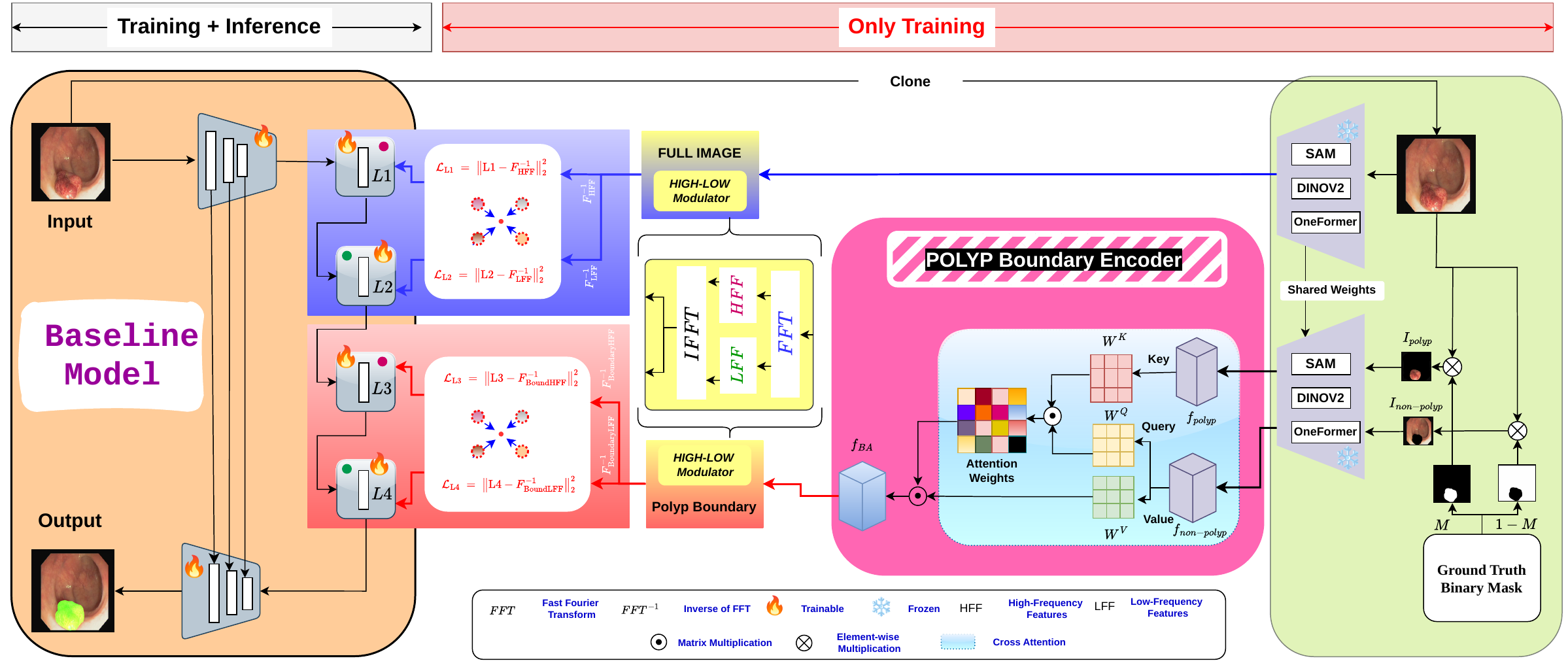}
   \vspace{-0.2cm}

   \caption{Overall architecture of the proposed \textbf{LiteBounD} framework. .} 

   \label{fig_model}
   \vspace{-0.5cm}
\end{figure*}

\section{Methodology}

We introduce \textbf{\underline{Li}gh\underline{t}w\underline{e}ight \underline{Boun}dary-guided \underline{D}istillation (LiteBounD)}, a modular framework that bridges the representational richness of Vision Foundation Models (VFMs) with the efficiency of lightweight segmentation backbones. LiteBounD distills complementary semantic and structural priors from SAM, DINOv2, and OneFormer into compact models such as U-Net and U-Net++, enabling precise polyp segmentation with minimal computational overhead. As illustrated in Fig.~\ref{fig_model}, LiteBounD orchestrates a multi-stage distillation pipeline that disentangles semantic and boundary cues, modulates them via frequency decomposition, and aligns them with latent representations in the lightweight baseline model.

\subsection{Baseline Model Architecture}

We redesign the encoder of a standard U-Net to produce disentangled latent representations that separately encode semantic and boundary-specific information. Given an input endoscopic image $I \in \mathbb{R}^{H \times W \times 3}$ (with $H = W = 352$), the encoder generates multi-scale feature maps as $I \rightarrow \{f_1, f_2, f_3, f_4\}$ where $f_i \in \mathbb{R}^{\frac{H}{2^i} \times \frac{W}{2^i} \times C_i}$ and $C_i \in \{64, 128, 256, 512\}.$ At the bottleneck layer, we extract four latent vectors: $\mathbf{L1}$: global semantic embedding via $1 \times 1$ convolution on $f_4$,  $\mathbf{L2}$: refined local semantic embedding via convolution on $\mathbf{L1}$, $\mathbf{L3}$: edge-sensitive boundary embedding via convolution on $\mathbf{L2}$, $\mathbf{L4}$: higher-order boundary context via convolution on $\mathbf{L3}$. These vectors are grouped as semantic pair $\{\mathbf{L1}, \mathbf{L2}\}$ and boundary-aware pair $\{\mathbf{L3}, \mathbf{L4}\}$, serving as alignment targets for distillation.

\subsection{Cross-Model Feature Extraction}

\paragraph{Semantic Feature Aggregation.}
We extract semantic embeddings from each VFM for the input image $I$. Let $F_i \in \mathbb{R}^{H_i \times W_i \times C_i}$ denote the feature map from the $i$-th model. All feature maps are resized to a common resolution $H' \times W'$ and concatenated to yield $f^{**} \in \mathbb{R}^{H' \times W' \times C^{*}}$:
\begin{equation}
\quad C^{*} = C_{\text{SAM}} + C_{\text{DINOv2}} + C_{\text{OneFormer}}.
\end{equation}
This unified semantic tensor captures complementary global context and serves as input to the frequency-aware distillation module.

\paragraph{Boundary-Aware Feature Extraction.}
To isolate boundary-sensitive cues, we construct region-specific inputs using the ground-truth mask $M \in \mathbb{R}^{H \times W}$:
\begin{equation}
I_{polyp} = M \otimes I, \quad I_{non-polyp} = (1 - M) \otimes I.
\end{equation}
These inputs are passed through each VFM to obtain polyp-aware and non-polyp-aware embeddings $f^{i}_{polyp}, f^{i}_{non-polyp} \in \mathbb{R}^{H_i \times W_i \times C_i}$.
After resizing and concatenation we get $f_{polyp}, f_{non-polyp} \in \mathbb{R}^{H' \times W' \times C^{*}}.$ We then apply cross-attention between $f_{polyp}$ (key) and $f_{non-polyp}$ (query/value) to generate boundary-enhanced features:
\begin{equation}
f_{BA} = \text{CrossAttn}(f_{polyp}, f_{non-polyp}).
\end{equation}

\subsection{Frequency-Aware Distillation}

To decompose features into structural and contextual components, we apply 2D FFT to $f^{**}$ and $f_{BA}$:
\begin{equation}
F^{\text{freq}} = \text{FFT}(f^{**}), \quad F^{\text{freq}}_{ba} = \text{FFT}(f_{BA}).
\end{equation}

Using binary masks $\text{MH}_{\text{HFF}}, \text{ML}_{\text{LFF}} \in \{0,1\}^{H' \times W'}$, we isolate High-Frequency Features (HFF) and Low-Frequency Features (LFF):
{\small\begin{equation}
\begin{aligned}
F^{-1}_{\text{LFF}} &= \text{IFFT}(\text{ML}_{\text{LFF}}, F^{\text{freq}}), F^{-1}_{\text{HFF}} &= \text{IFFT}(\text{MH}_{\text{HFF}}, F^{\text{freq}})\\
\end{aligned}
\end{equation}}
{\small
\begin{equation}
\begin{aligned}
F^{-1}_{\text{BoundLFF}} &= \text{IFFT}(\text{ML}_{\text{LFF}}, F^{\text{freq}}_{ba}), F^{-1}_{\text{BoundHFF}} &= \text{IFFT}(\text{MH}_{\text{HFF}}, F^{\text{freq}}_{ba})
\end{aligned}
\end{equation}
}

To enable feature-level distillation, these high- and low-frequency features are mapped back to the spatial domain using a 2D inverse FFT. The resulting distillation-ready features $F^{-1}_{\text{LFF}}$, $F^{-1}_{\text{HFF}}$, $F^{-1}_{\text{BoundLFF}}$, and $F^{-1}_{\text{BoundHFF}}$,
are then injected into the baseline to guide both semantic and boundary-guided learning.

\subsection{Latent Alignment Losses}

LiteBounD transfers VFM knowledge into the lightweight backbone through a dual-path latent alignment strategy that supervises semantic and structural representations separately. This design leverages the well-established observation that low-frequency features encode global semantics, while high-frequency features capture boundary-level details and rapid spatial variations. By aligning latent vectors with frequency-decomposed VFM features, LiteBounD enforces explicit representation disentanglement within the student model.

\paragraph{Semantic Alignment.}
Semantic alignment transfers global contextual priors and shape-level consistency from VFMs. The semantic latent vector $\mathbf{L1}$ is aligned with the low-frequency features $F^{-1}_{\text{LFF}}$ derived from the unified semantic embedding, while $\mathbf{L3}$ is aligned with $F^{-1}_{\text{BoundLFF}}$ extracted from boundary-aware features. These low-frequency signals encode dominant structural patterns and coarse object geometry, which are essential for stable polyp localization under varying imaging conditions.
{\small
\begin{equation}
\mathcal{L}_{\text{L1}} = \frac{1}{\text{HWC}^*} \sum_{x,y,z} \left\| L1(x,y,z) - F^{-1}_{\text{LFF}}(x,y,z) \right\|_2^2
\end{equation}
\begin{equation}
\mathcal{L}_{\text{L3}} = \frac{1}{\text{HWC}^*} \sum_{x,y,z} \left\| L3(x,y,z) - F^{-1}_{\text{BoundLFF}}(x,y,z) \right\|_2^2
\end{equation}
}

This supervision encourages the student model to internalize VFM-level semantic abstraction, improving robustness to polyp shape variability and reducing false negatives.

\paragraph{Structural Alignment.}
Structural alignment focuses on transferring fine-grained boundary cues that are critical for accurate delineation of polyp margins. The latent vectors $\mathbf{L2}$ and $\mathbf{L4}$ are aligned with high-frequency features $F^{-1}_{\text{HFF}}$ and $F^{-1}_{\text{BoundHFF}}$, respectively. These features emphasize sharp transitions and edge discontinuities, which strongly correlate with anatomical boundaries in endoscopic imagery.
{\small
\begin{equation}
\mathcal{L}_{\text{L2}} = \frac{1}{\text{HWC}^*} \sum_{x,y,z} \left\| L2(x,y,z) - F^{-1}_{\text{HFF}}(x,y,z) \right\|_2^2,
\end{equation}
\begin{equation}
\mathcal{L}_{\text{L4}} = \frac{1}{\text{HWC}^*} \sum_{x,y,z} \left\| L4(x,y,z) - F^{-1}_{\text{BoundHFF}}(x,y,z) \right\|_2^2
\end{equation}
}

By supervising boundary-aware latent vectors with high-frequency signals, LiteBounD enhances sensitivity to subtle contour variations and reduces over-smoothing—limitations commonly observed in CNN-based medical segmentation.

\textbf{Overall Impact.}  
The combination of semantic and structural alignment enables LiteBounD to jointly capture global coherence and boundary precision. This dual-path distillation allows lightweight models to approximate the representational richness of VFMs while retaining real-time efficiency. 

\subsection{Boundary-Aware Decoder}

The boundary-aware decoder reconstructs high-quality polyp masks by fusing multi-scale encoder features with distilled semantic and structural cues. It receives $\{f_1, f_2, f_3, f_4\}$ along with frequency-separated features, where low-frequency features $F^{-1}_{\text{LFF}}$ guide semantic refinement in stages ($L1$, $L3$), and high-frequency features $F^{-1}_{\text{HFF}}$ sharpen boundary-focused stages ($L2$, $L4$). This targeted fusion injects global contextual priors and boundary-sensitive structural information directly into the decoding pathway. By jointly enhancing semantic coherence and edge precision, the decoder produces accurate, pixel-level segmentation masks that effectively leverage the distilled knowledge from VFMs.

\definecolor{softblue}{RGB}{220,230,245}

\definecolor{softgreen}{RGB}{230,245,235}
\definecolor{softmint}{RGB}{235,250,240}
\definecolor{softolive}{RGB}{240,245,230}

\definecolor{softorange}{RGB}{255,240,225}
\definecolor{softpeach}{RGB}{255,235,220}
\definecolor{softsand}{RGB}{250,245,235}
\begin{table*}[ht]
\centering
\caption{Quantitative comparison of LiteBounD-guided baselines against state-of-the-art methods on seen datasets.}
\vspace{-0.2cm}
\label{tab:combined_seen}
\resizebox{\textwidth}{!}{%
\begin{tabular}{l|c|c|c|c|c|c|c|c|c|c|c|c|c|c}
\hline
\multirow{2}{*}{Methods} & \multirow{2}{*}{Params (M)} & \multirow{2}{*}{FLOPs (G)} & \multicolumn{6}{c|}{Kvasir-SEG (Seen)} & \multicolumn{6}{c}{CVC-ClinicDB (Seen)} \\ \cmidrule(lr){4-15} 
 & & & mDice ↑ & mIoU ↑ & $F_{\beta}^{w}$ ↑ & $S_{\alpha}$ ↑ & $E_{\phi}^{\text{max}}$ ↑ & MAE ↓ & mDice ↑ & mIoU ↑ & $F_{\beta}^{w}$ ↑ & $S_{\alpha}$ ↑ & $E_{\phi}^{\text{max}}$ ↑ & MAE ↓ \\ \midrule

\rowcolor{gray!15}
\multicolumn{15}{c}{\small \textbf{State-of-the-art Methods (Without Boundary Awareness)}} \\

SANet \cite{wei2021shallow} (MICCAI 2021)        & 23.8 & 11.3 & 90.4 & 84.7 & 89.2 & 91.5 & 95.3 & 2.8 & 91.6 & 85.9 & 90.9 & 93.9 & 97.6 & 1.2 \\ 
MSNet \cite{zhao2021automatic}  (MICCAI 2021)     & 27.6 & 17.0 & 90.7 & 86.2 & 89.3 & 92.2 & 94.4 & 2.8 & 92.1 & 87.9 & 91.4 & 94.1 & 97.2 & 0.8 \\
Polyp-PVT \cite{Dong2023}   (CAAI 2023)     & 25.1 & 10.1 & 91.7 & 86.4 & 91.1 & 92.5 & 95.6 & 2.3 & 93.7 & 88.9 & 93.6 & 94.9 & 98.5 & 0.6 \\
M$^2$SNet \cite{zhao2023m}  (arXiv 2023)       & 27.7 & 17.1 & 91.2 & 86.1 & 90.1 & 92.2 & 95.3 & 2.5 & 92.2 & 88.0 & 91.7 & 94.2 & 97.0 & 0.9 \\ 
PVT-Cascade \cite{rahman2023medical}  (WACV 2023)   & 35.2 & 32.5 & 91.1 & 86.3 & 90.6 & 91.9 & 96.1 & 2.5 & 91.9 & 87.2 & 91.8 & 93.6 & 96.9 & 1.3 \\ 
CTNet \cite{xiao2024ctnet} (IEEE TCYB 2024)         & 44.2 & 32.6 & 91.7 & 86.3 & 91.0 & 92.8 & 95.9 & 2.3 & 93.6 & 88.7 & 93.4 & 95.2 & 98.3 & 0.6 \\ 
SAM-Mamba \cite{dutta2025sam}  (WACV 2025)        & 103.0 & 423.0 & 92.4 & 87.3 & 94.2 & 93.6 & 96.1 & 2.5 & 94.2 & 88.7 & 94.3 & 95.5 & 98.2 & 0.6 \\ 
\midrule

\rowcolor{softblue}
\multicolumn{15}{c}{\small \textbf{State-of-the-art Methods (With Boundary Awareness)}} \\
\rowcolor{softblue} CFA-Net \cite{zhou2023cross}  (PR 2023)      & 25.2 & 55.3 & 91.5 & 86.1 & 90.3 & 92.4 & 96.2 & 2.3 & 93.3 & 88.3 & 92.4 & 95.0 & 98.9 & 0.7 \\ 
\rowcolor{softblue} MEGANet \cite{bui2024meganet}  (WACV 2024)     & 44.1 & 28.8 & 91.3 & 86.3 & 90.7 & 91.8 & 95.9 & 2.5 & 93.8 & 89.4 & 94.0 & 95.0 & 98.6 & 0.6 \\ 

\bottomrule

\rowcolor{softorange}
\multicolumn{15}{c}{\small \textbf{Lightweight Baseline Methods}} \\

\rowcolor{softorange} U-Net \cite{ronneberger2015u}  (MICCAI 2015)     & 16.7 & 73.9 & 81.8 & 74.6 & 79.4 & 85.8 & 89.3 & 5.5 & 82.3 & 75.5 & 81.1 & 88.9 & 95.4 & 1.9 \\
\rowcolor{softorange} U-Net++ \cite{zhou2018unet++}  (DLMIA 2018)     & 9.1 & 65.9 & 82.1 & 74.3 & 80.8 & 86.2 & 91.0 & 4.8 & 79.4 & 72.9 & 78.5 & 87.3 & 93.1 & 2.2 \\ 
\rowcolor{softorange} PraNet \cite{fan2020pranet}   (MICCAI 2020)    & 30.4 & 13.1 & 89.8 & 84.0 & 88.5 & 91.5 & 94.8 & 3.0 & 89.9 & 84.9 & 89.6 & 93.6 & 97.9 & 0.9 \\ 
\bottomrule

\rowcolor{softolive}
\multicolumn{15}{c}{\small \textbf{Lightweight Baselines Enhanced By Polyp-DiFoM \cite{agnihotri2026sam} (WACV 2026) }} \\

\rowcolor{softolive}
\textbf{U-Net + Polyp-DiFoM  }       & 16.9 & 74.7 & 86.6 & 76.4 & 85.3
& 94.1 & 91.0 & 4.1 & 93.9 & 88.7 & 92.7& 96.9 & 96.3 & 1.1 \\

\rowcolor{softolive}
\textbf{U-Net++ + Polyp-DiFoM}      & 9.6  & 66.4 & 84.7 & 74.7 & 83.3 & 93.7 & 92.9 & 4.8 & 89.5 & 83.4 & 91.1 & 96.7 & 95.4 & 1.9 \\

\rowcolor{softolive}
\textbf{PraNet + Polyp-DiFoM }      & 31.4 & 13.5 & 90.9 & 84.7 & 88.2 & 94.6 & 91.9 & 3.4 & 94.2 & 91.2 & 92.9 & 99.1 & 98.0 & 1.4 \\

\bottomrule

\rowcolor{green!45}
\multicolumn{15}{c}{\small \textbf{Lightweight Baselines Further Enhanced By \textcolor{purple}{LiteBounD (Ours)}}} \\

\rowcolor{green!45}
\textbf{U-Net + Ours }       & 17.0  & 73.1  & 86.9  & 77.6  & 86.1  & 94.3 
& 91.7 
& 3.9  
& 94.8 
& 89.7 
& 93.4 
& 97.6 
& 97.0 
& 0.9  \\

\rowcolor{green!45}
&
&
& \textcolor{purple}{(+5.1)}
& \textcolor{purple}{(+3.0)}
& \textcolor{purple}{(+6.7)}
& \textcolor{purple}{(+8.5)}
& \textcolor{purple}{(+2.4)}
& \textcolor{purple}{(-1.6)}
& \textcolor{purple}{(+12.5)}
& \textcolor{purple}{(+14.2)}
& \textcolor{purple}{(+12.3)}
& \textcolor{purple}{(+8.7)}
& \textcolor{purple}{(+1.6)}
&  \textcolor{purple}{(-1.0)} \\

\rowcolor{green!45}
\textbf{U-Net++ + Ours}      & 11.1  & 67.3  & 85.3  & 75.0  & 83.5 
& 93.9 
& 93.0 
& 4.7  
& 92.5 
& 87.8 
& 93.3 
& 98.2 
& 96.0 
& 1.6  \\

\rowcolor{green!45}
&
&
& \textcolor{purple}{(+3.2)}
& \textcolor{purple}{(+0.7)}
& \textcolor{purple}{(+2.7)}
& \textcolor{purple}{(+7.7)}
& \textcolor{purple}{(+2.0)}
& \textcolor{purple}{(-0.1)}
& \textcolor{purple}{(+13.1)}
& \textcolor{purple}{(+14.9)}
& \textcolor{purple}{(+14.8)}
& \textcolor{purple}{(+10.9)}
& \textcolor{purple}{(+2.9)}
& \textcolor{purple}{(-0.6)} \\

\rowcolor{green!45}
\textbf{PraNet + Ours }     & 32.8  & 14.8   & 91.9  & 84.1  & 88.3
& 94.1 
& 92.2 
& 3.2  
& 95.7 
& 91.3 
& 93.4 
& 99.0 
& 97.8 
& 1.1   \\

\rowcolor{green!45}
&
&
& \textcolor{purple}{(+2.1)}
& \textcolor{purple}{(+0.1)}
& (-0.2)
& \textcolor{purple}{(+2.6)}
& (-2.6)
& (+0.2)
& \textcolor{purple}{(+5.8)}
& \textcolor{purple}{(+6.4)}
& \textcolor{purple}{(+3.8)}
& \textcolor{purple}{(+5.4)}
& (-0.1)
& (+0.2) \\

\bottomrule

\end{tabular}%
}
\vspace{-0.3cm}
\end{table*}

\subsection{Multi-Phase Training}
\label{sec:training_strategy}

To effectively balance representation learning with knowledge transfer, we adopt a three phased training protocol which enables the model to progressively refine itself for robust and generalizable polyp segmentation.
\textbf{Phase I: Task-Specific Pre-training}
We first train a standard U-Net architecture from scratch using only segmentation supervision though combined segmentation loss:
\vspace{-0.25cm}
\begin{equation}
\mathcal{L}_{\text{phase1}} = \mathcal{L}_{\text{bce}} + \mathcal{L}_{\text{dice}} 
\end{equation}
where $\mathcal{L}_\text{bce}$ and $\mathcal{L}_\text{dice}$ denote the binary cross-entropy and dice losses, respectively.
\textbf{Phase II: Integrated Distillation}
In this phase, we activate the distillation modules, allowing the network to integrate both segmentation targets and rich semantic, boundary-aware priors extracted from the foundation models. The overall objective is expressed as:
{\small\begin{equation}
\mathcal{L}_{\text{phase2}}=\lambda_1 \left( \mathcal{L}_{\text{bce}} + \mathcal{L}_{\text{dice}} \right) + \lambda_2  \mathcal{L}_{\text{L1}}  + \lambda_3  \mathcal{L}_{\text{L2}} + \lambda_4  \mathcal{L}_{\text{L3}}  + \lambda_5  \mathcal{L}_{\text{L4}} 
\end{equation}}
where, we set $\lambda_1$ =
0.6 and $\lambda_2$ = $\lambda_3$ = $\lambda_4$ = $\lambda_5$ = 0.1 to prioritize semantic learning through segmentation while encouraging structural consistency via distillation.
\textbf{Phase III: Targeted Mask Distillation}
To stabilize the learned representations and increase decoding capability, we freeze the encoder in the final phase and optimize only the decoder and distillation pathways. The loss remains same as Phase~II, enabling focused refinement of mask while reducing overfitting.


\section{Experiments}

\subsection{Datasets and Performance Metrics}
\textbf{Datasets: }To evaluate the generalization and robustness of the proposed framework, we conduct experiments on five standard public benchmarks: Kvasir-SEG \cite{jha2019kvasir}, CVC-ClinicDB \cite{bernal2015wm}, ETIS \cite{silva2014toward}, CVC-ColonDB \cite{zhao2021automatic}, and EndoScene \cite{vazquez2017benchmark}. To ensure a fair comparison with recent state-of-the-art methods, we adopt the same data split protocol as in \cite{fan2020pranet}. In total, 1450 images (900 from Kvasir-SEG and 550 from CVC-ClinicDB) are used for training and the remaining 100 images from Kvasir-SEG and 62 images from CVC-ClinicDB are kept for testing. Additionally, to rigorously assess the generalization capability, three unseen datasets are considered: ETIS (196 images), CVC-ColonDB (380 images), and CVC-300 (60 images). \textbf{Metrics:} 
The performance of our model, as well as state-of-the-art methods, is quantitatively assessed using six standard metrics:
 the mean IoU (mIoU), mean Dice (mDice), Structure-measure ($S_{\alpha}$), weighted F-measure ($F^{w}_{\beta}$), Enhanced-alignment measure ($E^{\text{max}}_\phi$), and Mean Absolute Error (MAE). 
\subsection{Implementation Details}
The model is implemented using PyTorch framework on a NVIDIA Tesla V100 GPU (32 GB). All input images are rescaled to a resolution of $352\times352$ pixels. For augmentation, we apply random horizontal/vertical flipping and multi-scale resizing with factors $\{0.75, 1.0, 1.25\}$. Additionally, we perform training using the Adam optimizer with an initial learning rate of $1\times10^{-4}$ and default momentum parameters. In our work, training follows the three-phase strategy as described in Section~\ref{sec:training_strategy} for a total of 120 epochs (Phase I: Epochs 1–40, Phase II: Epochs 41–80 and Phase III: Epochs 81–120).


\section{Results Analysis}
\subsection{Quantitative Comparison}
To evaluate the effectiveness of the proposed LiteBounD, we choose three standard lightweight segmentation baselines: U-Net \cite{ronneberger2015u}, U-Net++ \cite{zhou2018unet++} and PraNet \cite{fan2020pranet}.

\definecolor{softblue}{RGB}{220,230,245}

\definecolor{softgreen}{RGB}{230,245,235}
\definecolor{softmint}{RGB}{235,250,240}
\definecolor{softolive}{RGB}{240,245,230}

\definecolor{softorange}{RGB}{255,240,225}
\definecolor{softpeach}{RGB}{255,235,220}
\definecolor{softsand}{RGB}{250,245,235}


\begin{table*}[ht]
\centering
\caption{Quantitative comparison of LiteBounD-guided baselines against state-of-the-art methods on unseen datasets. }
\vspace{-0.2cm}
\label{tab:combined_unseen}
\resizebox{\textwidth}{!}{%
\begin{tabular}{l|c|c|c|c|c|c|c|c|c|c|c|c|c|c|c|c|c|c}
\hline
\multirow{2}{*}{Methods} 
  & \multicolumn{6}{c|}{CVC-300 (Unseen)} 
  & \multicolumn{6}{c|}{CVC-ColonDB (Unseen)} 
  & \multicolumn{6}{c}{ETIS (Unseen)} \\ 
\cmidrule(lr){2-7}\cmidrule(lr){8-13}\cmidrule(lr){14-19}
 & mDice ↑ & mIoU ↑ & $F_{\beta}^{w}$ ↑ & $S_{\alpha}$ ↑ & $E_{\phi}^{\max}$ ↑ & MAE ↓ 
 & mDice ↑ & mIoU ↑ & $F_{\beta}^{w}$ ↑ & $S_{\alpha}$ ↑ & $E_{\phi}^{\max}$ ↑ & MAE ↓ 
 & mDice ↑ & mIoU ↑ & $F_{\beta}^{w}$ ↑ & $S_{\alpha}$ ↑ & $E_{\phi}^{\max}$ ↑ & MAE ↓ \\ 
\midrule

\rowcolor{gray!15}
\multicolumn{19}{c}{\small \textbf{State-of-the-art Methods (Without Awareness}} \\

SANet \cite{wei2021shallow}       
  & 88.8 & 81.5 & 85.9 & 92.8 & 97.2 & 0.8  
  & 75.3 & 67.0 & 72.6 & 83.7 & 87.8 & 4.3  
  & 75.0 & 65.4 & 68.5 & 84.9 & 89.7 & 1.5 \\

MSNet \cite{zhao2021automatic}    
  & 86.9 & 80.7 & 84.9 & 92.5 & 94.3 & 1.0  
  & 75.5 & 67.8 & 73.7 & 83.6 & 88.3 & 4.1  
  & 71.9 & 66.4 & 67.8 & 84.0 & 83.0 & 2.0 \\

Polyp-PVT \cite{Dong2023}  
  & 90.0 & 83.3 & 88.4 & 93.5 & 97.3 & 0.7  
  & 80.8 & 72.7 & 79.5 & 86.5 & 91.3 & 3.1  
  & 78.7 & 70.6 & 75.0 & 87.1 & 90.6 & 1.3 \\

M\textsuperscript{2}SNet \cite{zhao2023m}  
  & 90.3 & 84.2 & 88.1 & 93.9 & 96.5 & 0.9  
  & 75.8 & 68.5 & 73.7 & 84.2 & 86.9 & 3.8  
  & 74.9 & 67.8 & 71.2 & 84.6 & 87.2 & 1.7 \\

PVT-Cascade \cite{rahman2023medical}  
  & 89.2 & 82.4 & 87.3 & 93.2 & 95.9 & 0.9  
  & 78.1 & 71.0 & 77.9 & 85.5 & 89.6 & 3.1  
  & 78.6 & 71.2 & 75.9 & 87.2 & 89.6 & 1.3 \\

CTNet \cite{xiao2024ctnet}         
  & 90.8 & 84.4 & 89.4 & 97.5 & 97.5 & 0.6  
  & 81.3 & 73.4 & 80.1 & 87.4 & 91.5 & 2.7  
  & 81.0 & 73.4 & 77.6 & 88.6 & 91.3 & 1.4 \\

SAM-Mamba \cite{dutta2025sam}  
  & 92.0 & 86.1 & 88.8 & 94.6 & 98.1 & 0.6  
  & 85.3 & 77.1 & 85.6 & 89.8 & 93.3 & 1.7  
  & 84.8 & 78.2 & 85.5 & 91.6 & 93.3 & 1.0 \\ 
\midrule

\rowcolor{softblue}
\multicolumn{19}{c}{\small \textbf{State-of-the-art Methods (With Boundary Awareness)}} \\

\rowcolor{softblue} CFA-Net \cite{zhou2023cross}       
  & 89.3 & 82.7 & 93.8 & 87.5 & 97.8 & 0.8  
  & 74.3 & 66.5 & 72.8 & 83.5 & 89.8 & 3.9  
  & 73.2 & 65.5 & 69.3 & 84.5 & 89.2 & 1.4 \\

\rowcolor{softblue} MEGANet \cite{bui2024meganet}      
  & 89.9 & 83.4 & 88.2 & 93.5 & 96.9 & 0.7  
  & 79.3 & 71.4 & 77.9 & 85.4 & 89.5 & 4.0  
  & 73.9 & 66.5 & 70.2 & 83.6 & 85.8 & 3.7 \\

  \bottomrule

\rowcolor{softorange}
\multicolumn{19}{c}{\small \textbf{Lightweight Baseline Methods}} \\

\rowcolor{softorange} U-Net \cite{ronneberger2015u}      
  & 71.0 & 62.7 & 68.4 & 84.3 & 87.6 & 2.2  
  & 51.2 & 44.4 & 49.8 & 71.2 & 77.6 & 6.1  
  & 39.8 & 33.5 & 36.6 & 68.4 & 74.0 & 3.6 \\

\rowcolor{softorange} U-Net++ \cite{zhou2018unet++}      
  & 70.7 & 62.4 & 68.7 & 83.9 & 89.8 & 1.8  
  & 48.3 & 41.0 & 46.7 & 69.1 & 76.0 & 6.4  
  & 40.1 & 34.4 & 39.0 & 68.3 & 77.6 & 3.5 \\

\rowcolor{softorange} PraNet \cite{fan2020pranet}        
  & 87.1 & 79.7 & 84.3 & 92.5 & 97.2 & 1.0  
  & 70.9 & 64.0 & 69.6 & 81.9 & 86.9 & 4.5  
  & 62.8 & 56.7 & 60.0 & 79.4 & 84.1 & 3.1 \\

  \bottomrule

\rowcolor{softolive}
\multicolumn{19}{c}{\small \textbf{Lightweight Baselines Enhanced with Polyp-DiFoM \cite{agnihotri2026sam}}} \\

\rowcolor{softolive} \textbf{U-Net + PolypDiFoM}       
  & 82.3 & 73.3  & 74.7  & 89.4  & 86.9   & 1.6   
  & 68.3   & 57.4  & 60.9  & 81.4  & 77.1   & 4.7   
  & 54.3  & 42.4 & 47.1  & 76.6  & 70.1  & 2.6 \\

\rowcolor{softolive} \textbf{U-Net++ + PolypDiFoM}      
  & 77.8  & 70.9  & 75.9  & 92.7   & 90.1  & 1.5   
  & 67.7  & 54.1  & 64.0 & 87.1 & 80.0 & 5.1
  & 51.8  & 42.7  & 49.0 & 80.1  & 72.0   & 3.2 \\

\rowcolor{softolive} \textbf{PraNet + PolypDiFoM}       
  & 87.4  & 79.9  & 87.1  & 95.9  & 97.9   & 0.8  
  & 74.0  & 64.3  & 73.0  & 88.4  & 85.9 & 3.9  
  & 71.5  & 58.2  & 65.6  & 87.0   & 83.5   & 2.2 \\

\bottomrule

\rowcolor{green!45}
\multicolumn{19}{c}{\small \textbf{Lightweight Baselines Further Enhanced By \textcolor{purple}{LiteBounD}}} \\

\rowcolor{green!45} \textbf{U-Net + Ours}       
& 84.3 
& 75.1 
& 75.7 
& 91.3 
& 87.1 
& 1.3  
& 68.8 
& 58.2 
& 61.6 
& 81.6 
& 77.3
& 4.8  
& 54.9 
& 43.4 
& 48.2 
& 77.3 
& 72.9 
& 2.5   \\

\rowcolor{green!45}    
& \textcolor{purple}{(+13.3)}
& \textcolor{purple}{(+12.4)}
& \textcolor{purple}{(+7.3)}
& \textcolor{purple}{(+7.0)}
& (-0.5)
& \textcolor{purple}{(-0.9)}
& \textcolor{purple}{(+17.6)}
& \textcolor{purple}{(+13.8)}
& \textcolor{purple}{(+11.8)}
& \textcolor{purple}{(+10.4)}
& (-0.3)
& \textcolor{purple}{(-1.3)}
& \textcolor{purple}{(+15.1)}
& \textcolor{purple}{(+9.9)}
& \textcolor{purple}{(+11.6)}
& \textcolor{purple}{(+8.9)}
& (-1.1)
& \textcolor{purple}{(-1.1)} \\

\rowcolor{green!45} \textbf{U-Net++ + Ours}      
& 83.2 
& 73.3 
& 77.1 
& 93.3 
& 89.3
& 1.4 
& 68.9 
& 57.7 
& 65.1 
& 88.2
& 81.0 
& 4.8  
& 52.7 
& 43.0 
& 50.1 
& 80.7 
& 75.6 
& 3.2 \\

\rowcolor{green!45}     
& \textcolor{purple}{(+12.5)}
& \textcolor{purple}{(+10.9)}
& \textcolor{purple}{(+8.4)}
& \textcolor{purple}{(+9.4)}
& (-0.5)
& \textcolor{purple}{(-0.4)}
& \textcolor{purple}{(+20.6)}
& \textcolor{purple}{(+16.7)}
& \textcolor{purple}{(+18.4)}
& \textcolor{purple}{(+19.1)}
& \textcolor{purple}{(+5.0)}
& \textcolor{purple}{(-1.6)}
& \textcolor{purple}{(+12.6)}
& \textcolor{purple}{(+8.6)}
& \textcolor{purple}{(+11.1)}
& \textcolor{purple}{(+12.4)}
& (-2.0)
& \textcolor{purple}{(-0.3)} \\

\rowcolor{green!45} \textbf{PraNet + Ours}       
& 89.7 
& 81.7 
& 87.7 
& 96.5 
& 97.3 
& 0.7  
& 75.0 
& 66.7 
& 71.0 
& 90.3 
& 86.6 
& 4.1  
& 71.2 
& 60.1 
& 67.1 
& 88.3 
& 84.3 
& 2.0   \\

\rowcolor{green!45}    
& \textcolor{purple}{(+2.6)}
& \textcolor{purple}{(+2.0)}
& \textcolor{purple}{(+3.4)}
& \textcolor{purple}{(+4.0)}
& \textcolor{purple}{(+0.1)}
& \textcolor{purple}{(-0.3)}
& \textcolor{purple}{(+4.1)}
& \textcolor{purple}{(+2.7)}
& \textcolor{purple}{(+1.4)}
& \textcolor{purple}{(+8.4)}
& (-0.3)
& \textcolor{purple}{(-0.4)}
& \textcolor{purple}{(+8.4)}
& \textcolor{purple}{(+3.4)}
& \textcolor{purple}{(+7.1)}
& \textcolor{purple}{(+8.9)}
& \textcolor{purple}{(+0.2)}
& \textcolor{purple}{(-1.1)} \\

\bottomrule
\end{tabular}%
}
\vspace{-0.5cm}
\end{table*}

\textbf{Performance on Seen Datasets:}
Results on the seen datasets (Kvasir-SEG and CVC-ClinicDB) are summarized in Table~\ref{tab:combined_seen}. Our framework consistently outperforms all baseline architectures with minimal additional parameters and FLOPs. For the U-Net backbone, the Dice score improves by +5.1\% (81.8\% → 86.9\%) on Kvasir-SEG and +12.5\% (79.4\% → 94.8\%) on CVC-ClinicDB, with only a 0.3M parameter increase (16.7M → 17.0M). For U-Net++, it achieves gains of +3.2\% and +13.1\% on Kvasir-SEG and CVC-ClinicDB, respectively. For PraNet, it yields 95.7\% Dice and 91.3\% mIoU on CVC-ClinicDB, surpassing existing SOTA methods while maintaining high efficiency. Further, we compare LiteBounD with Polyp-DiFoM \cite{agnihotri2026sam} under a fair three-foundation-model (3F) setting. As shown in Table~\ref{tab:combined_seen}, LiteBounD consistently outperforms Polyp-DiFoM, achieving +3\% Dice and +4.4\% mIoU gains on CVC-ClinicDB (U-Net++) and +1\% Dice on Kvasir-SEG (PraNet). These results highlight the effectiveness of boundary-guided distillation in improving performance with fewer VFMs.

\textbf{Performance on Unseen Datasets:}
On unseen datasets (CVC-300, CVC-ColonDB, and ETIS), as shown in Table~\ref{tab:combined_unseen}, U-Net demonstrates substantial improvements, particularly on CVC-300, where the Dice score increases from 71.0\% to 84.3\% (+13.3\%) and mIoU from 62.7\% to 75.1\% (+12.4\%). Similar gains are observed on other challenging datasets, highlighting enhanced generalization capability and the effectiveness of learning robust, transferable features. In addition, LiteBounD consistently outperforms PolypDiFoM, with notable gains such as +5.4\% Dice on CVC-300 (U-Net++) and +2.4\% mIoU for PraNet. These improvements highlight the effectiveness of our model in enhancing generalization.


\begin{figure}[ht]
  \centering

  \begin{subfigure}[t]{\columnwidth}
    \centering
    \includegraphics[width=\linewidth,height=0.2\textheight]{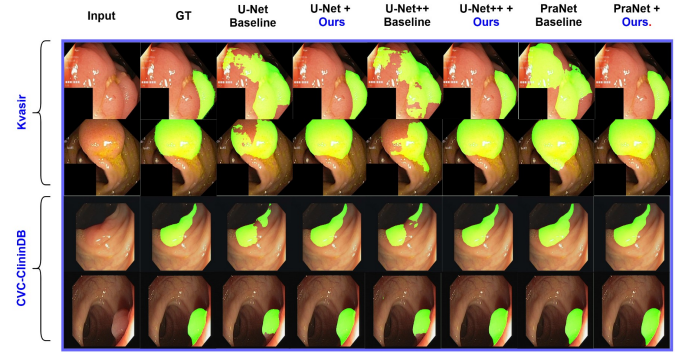}
    \caption{Seen datasets: Kvasir-SEG and CVC-ClinicDB.}
    \label{fig:qual_seen}
  \end{subfigure}
  
  \begin{subfigure}[t]{\columnwidth}
    \centering
    \includegraphics[width=\linewidth,height=0.28\textheight]{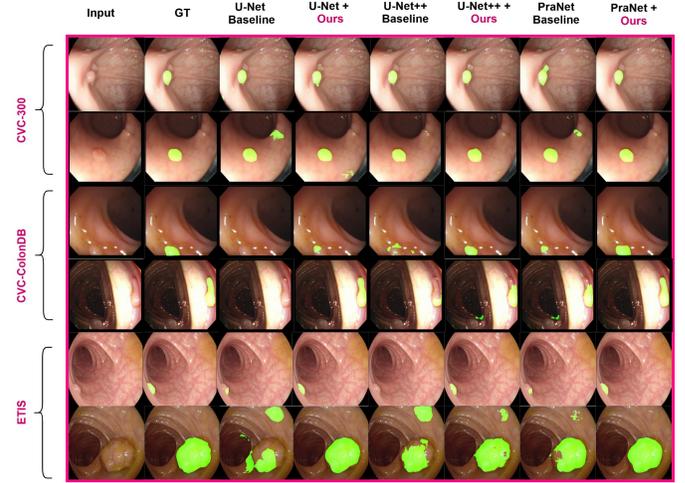}
    \caption{Unseen datasets: CVC-300, CVC-ColonDB, and ETIS.}
    \label{fig:qual_unseen}
  \end{subfigure}

  \caption{Qualitative comparison on seen and unseen datasets.} 
  \label{fig:qualitative}
  \vspace{-0.4cm}
\end{figure}
\subsection{Qualitative Comparison}
To further assess the effectiveness of our approach, we present qualitative comparisons between baseline models and their LiteBounD-enhanced counterparts across both seen and unseen datasets. As illustrated in Fig. \ref{fig:qualitative} baseline architectures often under-perform in challenging scenarios—such as weak boundaries, small-scale polyps, or low-contrast regions, where predicted masks either miss subtle structures or include spurious edge artifacts. In contrast, LiteBounD variants consistently produce sharper boundaries and more complete polyp delineations. 
For instance, the U-Net baseline exhibits highly inconsistent behavior: in some cases, it completely fails to detect the polyp, while in others it incorrectly segments non-polyp regions as foreground. In contrast, our model delivers substantially improved segmentation accuracy, producing more consistent and reliable predictions across challenging samples. U-Net++ and PraNet can detect polyps but often produce coarse boundaries while our boundary-aware mechanism refines edge representations, producing sharp, precise, and anatomically accurate polyp boundaries.

\subsection{Ablation Study}
Polyp-DiFoM~\cite{agnihotri2026sam} has previously shown that progressively incorporating task-specific and semantically complementary foundation models leads to consistent improvements in segmentation performance. Specifically, the 1F (SAM only) and 2F (SAM + DINOv2) settings were shown to yield comparatively limited gains, indicating that richer multi-model supervision is necessary to fully exploit the benefits of distillation. Motivated by these findings, we directly adopt the 3F setting (SAM, DINOv2, and OneFormer) as our baseline distillation configuration, avoiding the less effective 1F and 2F variants. The primary objective of this ablation is to examine the role of boundary awareness in challenging polyp segmentation scenarios. We perform experiments with and without the Boundary-Guided Distillation mechanism. As reported in Table \ref{tab:ablation}, the inclusion of mechanism consistently leads to improved performance, yielding notable gains in both Dice and IoU scores across baselines. These results confirm the effectiveness of explicitly modeling boundary information. 











\begin{table}[t]
\small
\centering
\caption{Effectiveness of Boundary-Guided Distillation}
\vspace{-0.1cm}
\resizebox{\columnwidth}{!}{
\begin{tabular}{c|cc|ccc}
\toprule
& \multicolumn{2}{c|}{\bf \small Seen} & \multicolumn{3}{c}{\bf \small Unseen} \\

 \textbf{Boundary-Guidance} 
& \bf \small Kvasir & \bf \small CVC-ClinicDB 
& \bf \small CVC-300 & \bf \small CVC-ColonDB & \bf \small ETIS \\ 
\midrule

\rowcolor{red!15}
\multicolumn{6}{c}{\textbf{\textcolor{blue}{U-Net}}} \\

\xmark & 86.6 & 93.9 & 82.3 & 68.3 & 54.3 \\
\checkmark & \textbf{86.9} & \textbf{94.8} & \textbf{84.3} & \textbf{68.8} & \textbf{54.9} \\ 
\midrule

\rowcolor{red!15}
\multicolumn{6}{c}{\textbf{\textcolor{blue}{U-Net++}}} \\

 \xmark & \textbf{84.7} & 89.5 & 77.8  & 67.7  & 51.8  \\
\checkmark &  84.1 & \textbf{92.5} & \textbf{83.2} & \textbf{68.9} & \textbf{52.7} \\ 
\midrule

\rowcolor{red!15}
\multicolumn{6}{c}{\textbf{\textcolor{blue}{PraNet}}} \\

\xmark & 90.9 & 94.2 & 87.4  & 74.0 & \textbf{71.5} \\
\checkmark & \textbf{91.9} & \textbf{95.7} & \textbf{89.7} & \textbf{75.0} & 71.2 \\ 
\bottomrule
\end{tabular}
}\vspace{-0.5cm}
\label{tab:ablation}
\end{table}

\section{Conclusion}
This paper introduces LiteBounD, a novel and effective boundary-guided distillation framework that leverages the rich representational capacity of large-scale vision foundation models to boost the performance of lightweight medical image segmentation architectures. By systematically generating strong semantic features from foundation models, converting them into rich boundary-aware representations, and fusing them into lightweight baselines,  LiteBounD substantially enhances the accuracy, robustness, and generalizability of these baselines, while preserving their computational efficiency. Extensive evaluations conducted on five widely used polyp segmentation benchmarks demonstrate that LiteBounD achieves comparable or better performance than recent state-of-the-art methods with significantly lower computational overhead. Overall, LiteBounD enables real-time polyp segmentation with high precision in weak-boundary cases, without heavy computation.

\bibliographystyle{IEEEtran}
\bibliography{refs}

@String(ICIP = {IEEE Int. Conf. Image Process.})

@String(ICIP  = {ICIP})

@inproceedings{fang2019selective,
  title={Selective feature aggregation network with area-boundary constraints for polyp segmentation},
  author={Fang, Yuqi and Chen, Cheng and Yuan, Yixuan and Tong, Kai-yu},
  booktitle={International Conference on Medical Image Computing and Computer-Assisted Intervention},
  pages={302--310},
  year={2019},
  organization={Springer}
}

@inproceedings{zhao2021automatic,
  title={Automatic polyp segmentation via multi-scale subtraction network},
  author={Zhao, Xiaoqi and Zhang, Lihe and Lu, Huchuan},
  booktitle={International Conference on Medical Image Computing and Computer-Assisted Intervention},
  pages={120--130},
  year={2021},
  organization={Springer}
}

@inproceedings{ronneberger2015u,
  title={U-net: Convolutional networks for biomedical image segmentation},
  author={Ronneberger, Olaf and Fischer, Philipp and Brox, Thomas},
  booktitle={International Conference on Medical Image Computing and Computer-Assisted Intervention},
  pages={234--241},
  year={2015},
  organization={Springer}
}

@inproceedings{fan2020pranet,
  title={Pranet: Parallel reverse attention network for polyp segmentation},
  author={Fan, Deng-Ping and Ji, Ge-Peng and others},
  booktitle={International Conference on Medical Image Computing and Computer-Assisted Intervention},
  pages={263--273},
  year={2020},
  organization={Springer}
}

@article{morgan2023global,
  title={Global burden of colorectal cancer in 2020 and 2040: incidence and mortality estimates from GLOBOCAN},
  author={Morgan, Eileen and Arnold, Melina and others},
  journal={Gut},
  volume={72},
  number={2},
  pages={338--344},
  year={2023},
  publisher={BMJ Publishing Group}
}

@inproceedings{jain2023oneformer,
  title={Oneformer: One transformer to rule universal image segmentation},
  author={Jain, Jitesh and Li, Jiachen and Chiu, Mang Tik and others},
  booktitle={Proceedings of the IEEE/CVF Conference on Computer Vision and Pattern Recognition},
  pages={2989--2998},
  year={2023}
}

@inproceedings{bui2024meganet,
  title={Meganet: Multi-scale edge-guided attention network for weak boundary polyp segmentation},
  author={Bui, Nhat-Tan and Hoang, Dinh-Hieu and others},
  booktitle={Proceedings of the IEEE/CVF Winter Conference on Applications of Computer Vision},
  pages={7985--7994},
  year={2024}
}

@article{xiao2024ctnet,
  title={Ctnet: Contrastive transformer network for polyp segmentation},
  author={Xiao, Bin and Hu, Jinwu and Li, Weisheng and Pun, Chi-Man and Bi, Xiuli},
  journal={IEEE Transactions on Cybernetics},
  volume={54},
  number={9},
  pages={5040--5053},
  year={2024},
  publisher={IEEE}
}

@inproceedings{rahman2023medical,
  title={Medical image segmentation via cascaded attention decoding},
  author={Rahman, Md Mostafijur and Marculescu, Radu},
  booktitle={Proceedings of the IEEE/CVF Winter Conference on Applications of Computer Vision},
  pages={6222--6231},
  year={2023}
}

@inproceedings{zhou2018unet++,
  title={Unet++: A nested u-net architecture for medical image segmentation},
  author={Zhou, Zongwei and Rahman Siddiquee, Md Mahfuzur and Tajbakhsh, Nima and Liang, Jianming},
  booktitle={International Workshop on Deep Learning in Medical Image Analysis},
  pages={3--11},
  year={2018},
  organization={Springer}
}

@inproceedings{kirillov2023segment,
  title={Segment anything},
  author={Kirillov, Alexander and Mintun, Eric and others},
  booktitle={Proceedings of the IEEE/CVF International Conference on Computer Vision},
  pages={4015--4026},
  year={2023}
}

@inproceedings{radford2021learning,
  title={Learning transferable visual models from natural language supervision},
  author={Radford, Alec and Kim, Jong Wook and others},
  booktitle={International Conference on Machine Learning},
  pages={8748--8763},
  year={2021},
  organization={PmLR}
}

@inproceedings{dutta2025sam,
  title={SAM-Mamba: Mamba Guided SAM Architecture for Generalized Zero-Shot Polyp Segmentation},
  author={Dutta, Tapas Kumar and Majhi, Snehashis and Nayak, Deepak Ranjan and Jha, Debesh},
  booktitle={2025 IEEE/CVF Winter Conference on Applications of Computer Vision (WACV)},
  pages={4655--4664},
  year={2025},
  organization={IEEE}
}

@article{silva2014toward,
  title={Toward embedded detection of polyps in wce images for early diagnosis of colorectal cancer},
  author={Silva, Juan and Histace, Aymeric and others},
  journal={International Journal of Computer Assisted Radiology and Surgery},
  volume={9},
  number={2},
  pages={283--293},
  year={2014},
  publisher={Springer}
}

@article{vazquez2017benchmark,
  title={A benchmark for endoluminal scene segmentation of colonoscopy images},
  author={V{\'a}zquez, David and Bernal, Jorge and others},
  journal={Journal of Healthcare Engineering},
  volume={2017},
  number={1},
  pages={4037190},
  year={2017},
  publisher={Wiley Online Library}
}

@inproceedings{jha2019kvasir,
  title={Kvasir-seg: A segmented polyp dataset},
  author={Jha, Debesh and Smedsrud, Pia H and others},
  booktitle={International Conference on Multimedia Modeling},
  pages={451--462},
  year={2019},
  organization={Springer}
}

@article{bernal2015wm,
  title={WM-DOVA maps for accurate polyp highlighting in colonoscopy: Validation vs. saliency maps from physicians},
  author={Bernal, Jorge and S{\'a}nchez, F Javier and others},
  journal={Computerized Medical Imaging and Graphics},
  volume={43},
  pages={99--111},
  year={2015},
  publisher={Elsevier}
}

@inproceedings{wei2021shallow,
  title={Shallow attention network for polyp segmentation},
  author={Wei, Jun and Hu, Yiwen and Zhang, Ruimao and Li, Zhen and Zhou, S Kevin and Cui, Shuguang},
  booktitle={International Conference on Medical Image Computing and Computer-Assisted Intervention},
  pages={699--708},
  year={2021},
  organization={Springer}
}

@article{zhou2023cross,
  title={Cross-level feature aggregation network for polyp segmentation},
  author={Zhou, Tao and Zhou, Yi and He, Kelei and Gong, Chen and Yang, Jian and Fu, Huazhu and Shen, Dinggang},
  journal={Pattern Recognition},
  volume={140},
  pages={109555},
  year={2023},
  publisher={Elsevier}
}

@article{zhao2023m,
  title = {M$^{2}$ SNet: Multi-scale in multi-scale subtraction network for medical image segmentation},
  author={Zhao, Xiaoqi and Jia, Hongpeng and Pang, Youwei and Lv, Long and Tian, Feng and Zhang, Lihe and Sun, Weibing and Lu, Huchuan},
  journal={arXiv preprint arXiv:2303.10894},
  year={2023}
}

@article{ahn2012miss,
  title={The miss rate for colorectal adenoma determined by quality-adjusted, back-to-back colonoscopies},
  author={Ahn, Sang Bong and Han, Dong Soo and Bae, Joong Ho and Byun, Tae Jun and Kim, Jong Pyo and Eun, Chang Soo},
  journal={Gut and liver},
  volume={6},
  number={1},
  pages={64},
  year={2012}
}

@inproceedings{cheng2022masked,
  title={Masked-attention mask transformer for universal image segmentation},
  author={Cheng, Bowen and Misra, Ishan and Schwing, Alexander G and others},
  booktitle={Proceedings of the IEEE/CVF conference on Computer Vision and Pattern Recognition},
  pages={1290--1299},
  year={2022}
}

@article{Dong2023, 
author = {Bo Dong and Wenhai Wang and Deng-Ping Fan and Jinpeng Li and Huazhu Fu and Ling Shao},
title = {Polyp-PVT: Polyp Segmentation with Pyramid Vision Transformers},
year = {2023},
journal = {CAAI Artificial Intelligence Research},
volume = {2},
pages = {9150015},
doi = {10.26599/AIR.2023.9150015},
}

@article{
oquab2024dinov,
title={{DINO}v2: Learning Robust Visual Features without Supervision},
author={Maxime Oquab and Timoth{\'e}e Darcet and Th{\'e}o Moutakanni and others},
journal={Transactions on Machine Learning Research},
issn={2835-8856},
year={2024},
note={Featured Certification}
}

@article{cheng2021per,
  title={Per-pixel classification is not all you need for semantic segmentation},
  author={Cheng, Bowen and Schwing, Alex and Kirillov, Alexander},
  journal={Advances in Neural Information Processing Systems},
  volume={34},
  pages={17864--17875},
  year={2021}
}

@inproceedings{agnihotri2026sam,
  title={From SAM to DINOv2: Towards Distilling Foundation Models to Lightweight Baselines for Generalized Polyp Segmentation},
  author={Agnihotri, Shivanshu and Majhi, Snehashis and Nayak, Deepak Ranjan and Jha, Debesh},
  booktitle={Proceedings of the IEEE/CVF Winter Conference on Applications of Computer Vision},
  pages={1757--1766},
  year={2026}
}

@inproceedings{jha2019resunet++,
  title={Resunet++: An advanced architecture for medical image segmentation},
  author={Jha, Debesh and Smedsrud, Pia H and Riegler, Michael A and Johansen, Dag and De Lange, Thomas and Halvorsen, P{\aa}l and Johansen, H{\aa}vard D},
  booktitle={2019 IEEE international symposium on multimedia (ISM)},
  pages={225--2255},
  year={2019},
  organization={IEEE}
}

@inproceedings{chakraborti2024mct,
  title={MCT-Net: a lightweight multiscale convolutional transformer network for polyp segmentation},
  author={Chakraborti, Niladri and Nayak, Deepak Ranjan},
  booktitle={2024 IEEE International Conference on Image Processing (ICIP)},
  pages={2944--2950},
  year={2024},
  organization={IEEE}
}

\end{document}